\documentclass{article}
\usepackage[final]{neurips_2024}

\usepackage[utf8]{inputenc} 
\usepackage[T1]{fontenc}    
\usepackage{hyperref}       
\usepackage{url}            
\usepackage{booktabs}       
\usepackage{amsfonts}       
\usepackage{nicefrac}       
\usepackage{microtype}      
\usepackage{xcolor}         

\setcitestyle{numbers,square,comma}

\title{Causal Abstraction in Model Interpretability: \\A Compact Survey}

\author{%
  Yihao Zhang \\
  School of Mathematical Sciences, Peking University\\
  \texttt{zhangyihao@stu.pku.edu.cn} \\
}

\begin{document}

\maketitle

\begin{abstract}
  The pursuit of interpretable artificial intelligence has led to significant advancements in the development of methods that aim to explain the decision-making processes of complex models, such as deep learning systems. Among these methods, causal abstraction stands out as a theoretical framework that provides a principled approach to understanding and explaining the causal mechanisms underlying model behavior. This survey paper delves into the realm of causal abstraction, examining its theoretical foundations, practical applications, and implications for the field of model interpretability.
\end{abstract}

\section{Introduction}

The increasing complexity of modern machine learning models, particularly deep learning systems, has raised concerns about their interpretability and transparency \cite{bereska2024mechanisticinterpretabilityaisafety}. As these models are deployed in a wide range of applications, from healthcare to finance, it is crucial to understand how they make decisions and to ensure that their outputs are reliable and trustworthy. The field of model interpretability has thus emerged as a critical area of research, with the goal of developing methods that can explain the decision-making processes of complex models in a human-understandable manner.

Recent research on the interpretability of complex systems has increasingly focused on understanding what drives specific behaviors in machine learning models—particularly black-box models like neural networks. These models, due to their vast number of parameters and complex architectures, often make it difficult for humans to discern how decisions are made. One promising approach to tackling this challenge is \textbf{causal abstraction} \cite{rubenstein2017causal,beckers2019abstracting,geiger2024causalabstractiontheoreticalfoundation}, which provides a formal framework for tracing the specific causes of model behaviors. Rather than just simplifying or approximating a model, causal abstraction enables researchers to examine the internal variables and mechanisms responsible for generating certain outputs, offering a clearer window into how the model operates at a deeper, mechanistic level.

This concept has its origins in fields like program analysis, where abstraction is used to isolate the critical components of complex systems \cite{10.1145/234528.234740}. However, causal abstraction goes beyond simplifying a model—it systematically identifies and traces causal pathways within the model, allowing for precise explanations of what caused certain behaviors. This approach aligns with the growing interest in \textit{mechanistic interpretability}, where the goal is to understand the internal workings of a model by isolating and examining the variables and mechanisms that directly influence its behavior. By revealing these causal links, causal abstraction improves our ability to interpret, trust, and explain machine learning models in a more transparent and accountable way.

In this survey paper, we delve into the realm of causal abstraction, exploring its theoretical foundations, practical applications, and implications for the field of model interpretability. Based on two newest research on causal abstraction for model explanation \cite{10.5555/3666122.3669540,geiger2024causalabstractiontheoreticalfoundation}, we begin by introducing the development of the concept of causal abstraction and its possible relationship to problems in causal reasoning and causal inference. We then discuss how causal abstraction can be applied to interpret machine learning models, focusing on its use in explaining the decision-making processes of complex models. Finally, we examine the challenges and opportunities in the field of causal abstraction and discuss future directions for research in this area.

\section{Development of Causal Abstraction Theory}

The concept of \textbf{causal abstraction} draws foundational support from the work of Rubenstein et al. in their 2017 paper, \textit{Causal Consistency of Structural Equation Models}~\cite{rubenstein2017causal}. In this paper, they proposed a theoretical framework of \textbf{exact transformations}, which formalizes the conditions under which two models are equivalent in a causal sense. The framework focuses on ensuring that causal mechanisms are preserved between models, addressing the issue of causal consistency.

Their motivating example came from a model linking total cholesterol (TC) to heart disease (HD), where experimental results paradoxically indicated that higher TC could both increase and decrease HD risk. Modern understanding, which differentiates between two types of cholesterol—HDL and LDL—each with distinct effects on HD, showed that the original model broke causal consistency. Rubenstein et al.'s "exact transformations" provided a formal method to determine when two models remain consistent in terms of their causal structure. 

While Rubenstein et al. did not introduce the idea of causal abstraction directly, their work laid the theoretical groundwork by offering a formal mathematical basis for determining causal equivalence between models. This provides a critical foundation for later work in causal abstraction, which uses these principles to construct simplified models that preserve key causal relationships, enabling more interpretable model explanations.

Causal abstraction was formally introduced by Beckers et al. in 2018 \cite{beckers2019abstracting}, building on Halpern’s foundational discussions on causal models \cite{10.7551/mitpress/10809.001.0001}. In this framework, structural equation models are formalized using a \textit{signature}, which consists of a set of variables and their associated possible values. The causal relationships between these variables—referred to as \textbf{mechanisms} in later works—are represented by a set of functions. Each function determines the value of an endogenous variable based on the values of other variables. \textit{Interventions} are formalized by replacing the mechanisms of certain variables with fixed values, a process analogous to \textit{do-calculus} \cite{pearl1994probabilistic}, and are referred to as hard interventions in subsequent literature. Building on these formalized structures, causal abstraction is initially defined as an equivalence relationship between a low-level model and a high-level model, where all interventions in the low-level model can be mapped to the high-level model via a function that satisfies specific properties. This framework provides a series of progressively more restrictive definitions of abstraction for causal models.

In their paper, Beckers et al. take an important step by demonstrating the possibility of abstracting a complex system of variables into a simpler one through the aggregation of variables into \textit{macro-variables}, while preserving the underlying causal mechanisms. This abstraction is critical for the development of causal modeling, where simpler models can still capture the essential causal relationships of more complex systems. 

This theory has been further developed in subsequent works. In 2020, Beckers et al. extended the concept of causal abstraction to account for more realistic scenarios where an abstract causal model serves as an approximation of the underlying system \cite{beckers2020approximate}. In this extended framework, they address the discrepancies that may arise between low- and high-level causal models of the same system, and provide a formal account of how one causal model can approximate another—a topic of independent interest.

The framework introduces the notion of \textit{approximate casual abstraction}, which is extended to probabilistic causal models, allowing for uncertainty in the abstracted model. To quantify the approximation, a distance function is used on certain features or variables to measure the difference between two models. This approach can be naturally extended to probabilistic causal models by considering the expected value of the distance function, where the hidden exogenous variables follow some probabilistic distribution. This work broadens the scope of causal abstraction, providing a structured method to account for imperfections and uncertainty in causal modeling.

By 2022, the foundations of causal abstraction theory were further solidified. Otsuka et al.~\cite{otsuka2022equivalence} extended the theory by leveraging the mathematical framework of category theory. They developed a category-theoretic criterion for determining the equivalence of causal models that possess different, yet homomorphic, directed acyclic graphs over discrete variables. A causal model, in their formulation, is interpreted probabilistically as a causal string diagram following the definition by Jacobs et al.~\cite{jacobs2019causal}. Specifically, it is defined as a functor from the “syntactic” category \textbf{Syn}$_G$, which represents the causal structure of a graph \textit{G}, to the category \textbf{Stoch}, consisting of finite sets and stochastic matrices. The equivalence of causal models is formally characterized through a natural transformation or isomorphism between these functors, which they term $\Phi$-abstraction and $\Phi$-equivalence, respectively. This work provides a rigorous and formal foundation for the concept of causal abstraction, establishing precise criteria for determining when two causal models can be considered equivalent.

In a related development, Zennaro~\cite{zennaro2022abstraction} reviewed the various approaches to abstracting equivalent causal models proposed to date, with a focus on the formal properties of mappings between structural causal models. Zennaro’s analysis highlighted the different layers—structural and distributional—at which these properties can be enforced, allowing for the distinction between families of abstractions based on which properties are guaranteed or prioritized.

In 2023, Massidda et al.~\cite{pmlr-v213-massidda23a} further generalized interventions in causal models by introducing the concept of \textit{soft interventions}. While previous research incorporated interventions similar to those used in do-calculus, such interventions were typically formulated as $do(x)$, which simply assigns a fixed value to the selected variables. However, in many cases, interventions are not limited to assigning specific values. Instead, we may wish to modify the underlying causal mechanisms themselves as part of the intervention.

A notable example arises when low-level causal models are abstracted into high-level models. In this scenario, interventions on low-level variables may only affect certain variables within a clustered macro-variable. When dealing with macro-variables, the intervention cannot be straightforwardly interpreted as a hard intervention. This discrepancy complicates the commutativity of diagrams in intervened causal models across different levels of abstraction. To address this, soft interventions and the corresponding notion of \textit{soft abstraction} were introduced, thereby extending the boundaries of causal abstraction to account for these more flexible forms of intervention.

Another significant advancement in the field of causal abstraction comes from the work of Geiger et al. \cite{geiger2020neural,10.5555/3540261.3540994} called \textit{interchange intervention}, which was further formalized in their subsequent paper \cite{geiger2024causalabstractiontheoreticalfoundation}. Their approach focuses on the application of causal abstraction techniques to neural networks. The motivation for this idea stems from investigating how a language model stores and utilizes information related to lexical entailment. By swapping a vector generated at a particular layer of the model from one input with a vector from another input, they explore the causal influence of internal model representations on the model's final behavior.

This method allows for the calculation of potential causal relationships between the internal vectors of the model and its output behavior, thereby providing a means to trace causality from within the model to its observable behavior. This idea of analyzing causal effects within models has been utilized in recent model interpretability related research as well \cite{chenselfie}. Geiger et al. formalized this process by defining a causal model that includes input and target variables, where the effects of different inputs on these variables can be observed and measured. This framework offers a new perspective on causal abstraction for models that explicitly depend on specific input values, making them ideal candidates for intervention-based causal analysis, as their mechanisms are not influenced by other variables.

To summarize, causal abstraction, as a formal framework, has evolved significantly from its early theoretical foundations in the work of Rubenstein et al., which established causal consistency through exact transformations, to more modern interpretations involving neural networks and complex interventions. Over time, the theory has been enriched by contributions that extend causal abstraction to account for uncertainty, probabilistic models, and soft interventions, all while ensuring that the core causal mechanisms are preserved. These developments have provided a robust structure for simplifying complex causal models into more interpretable forms, enabling researchers to address real-world challenges in areas ranging from artificial intelligence to medical research. As causal abstraction continues to evolve, it promises to enhance our ability to model, understand, and intervene in complex systems while maintaining a clear and mathematically rigorous foundation.

\section{Enhancement of Model Interpretability through Causal Abstraction}

Causal abstraction provides a formal framework for improving the interpretability of complex machine learning models by revealing the underlying causal mechanisms driving their behavior. Rather than merely simplifying a model, causal abstraction helps researchers pinpoint specific causes behind model outputs, offering insights that are more intuitive and aligned with human reasoning. This section highlights practical applications where causal abstraction has been employed to interpret models across various domains.

The first study to apply causal abstraction to artificial intelligence models for interpretability was conducted by Geiger et al. \cite{geiger2020neural}, who laid the initial theoretical foundation for causal abstraction in neural networks. While neural networks are often treated as black boxes, they have the distinct advantage of being highly observable, making them well-suited as a base model for causal abstraction. Geiger et al. introduced an initial method, \textit{causal abstraction analysis}, to evaluate neural networks by first positing hypotheses in the form of a causal model. They then assessed the alignment of neural representations with this hypothesized causal model using an algorithm called \textit{distributed alignment search}.

To validate whether the neural representations align with high-level causal variables, they employed the \textit{interchange intervention} method. This approach exemplifies how causal abstraction can be extended beyond theoretical frameworks to practical applications in model interpretation. By using interchange interventions, they experimentally assessed whether the causal model holds within the neural network, wherein the high-level model serves as an abstraction of the neural network’s mechanisms. Specifically, \textit{alignment search} calculates how closely the lower-level model aligns with the high-level causal model. Here, interchange interventions are critical, experimentally connecting internal representations to the causal model through counterfactual testing. Although abstraction itself is challenging, this counterfactual approach allows researchers to verify causal effects, bridging causal abstraction with practical model interpretability.

The methodology of Distributed Alignment Search (DAS) was further developed by Wu et al. in their 2023 work on interpretability at scale, particularly for large-scale language models \cite{10.5555/3666122.3669540}. In this study, they introduced Boundless DAS, an enhanced version of DAS designed to efficiently explore causal alignments in models with billions of parameters, such as the Alpaca model. Boundless DAS improves upon the original by replacing brute-force search steps with learnable parameters, thus enabling effective causal analysis across distributed neural representations.

In their work, Wu et al. applied Boundless DAS to the Alpaca model, investigating its causal mechanisms in a simple numerical reasoning task. By identifying causal alignments between interpretable Boolean variables and neural representations within Alpaca, they demonstrated that causal abstraction could scale robustly, retaining faithful alignment even under variations in input and context. This adaptation marks a significant step forward in applying causal abstraction to large language models, advancing model interpretability in a practical, scalable manner.

The other approach to causal abstraction in neural networks was proposed by Geiger et al. in their 2024 paper \cite{geiger2024causalabstractiontheoreticalfoundation}. They introduced a theoretical foundation for causal abstraction in neural networks, focusing on the interchange intervention method. This method allows for the calculation of potential causal relationships between internal vectors of the model and its output behavior, providing a means to trace causality from within the model to its observable behavior. By defining a causal model that includes input and target variables, they formalized the process of analyzing causal effects within models, offering a new perspective on causal abstraction for models that explicitly depend on specific input values.

We finally introduce the recent work by Geiger et al., which presents a unified framework for mechanistic interpretability by revisiting the foundational principles of causal abstraction and extending them to formalize a variety of interpretability methods for neural networks \cite{geiger2024causalabstractiontheoreticalfoundation}. This framework builds upon and enhances the concept of \textbf{mechanistic interpretability} \cite{bereska2024mechanisticinterpretabilityaisafety}, a crucial approach within AI interpretability focused on reverse-engineering the computational mechanisms and representations learned by neural networks. Mechanistic interpretability aims to translate these learned representations into human-understandable algorithms and concepts, providing a fine-grained, causal understanding that is essential for aligning AI systems with human values and ensuring safe operation. This emphasis underscores the increasing relevance of causality-based techniques in advancing mechanistic insights.

Developed from the foundations of causal abstraction, this work provides a more comprehensive framework by integrating several previous theoretical advances. In particular, hard and soft interventions are systematically combined within an algebraic structure, forming what Geiger et al. term an \textit{intervention algebra}. This algebra formalizes calculation rules that support both recursive and approximate applications of causal abstraction, enabling a more nuanced and flexible approach to interpretability in complex systems.

Additionally, the authors further extend interchange intervention theory to accommodate recursive and approximate scenarios, addressing the needs of more intricate abstraction applications. By generalizing causal abstraction theory from mechanism replacement—incorporating both hard and soft interventions—to arbitrary mechanism transformations, they introduce functional transformations that map original mechanisms to new ones. This expansion allows for a broader range of interpretability applications, as it enables the causal framework to support dynamic modifications across diverse, complex model structures. 

The primary theoretical contribution of this work lies in its provision of an exact methodology for translating existing concepts in mechanistic interpretability into the framework of causal abstraction, effectively establishing an isomorphic relationship between interpretability methods and causal abstraction techniques. This translation allows interpretability methods to be rigorously understood as instances of causal abstraction, thereby enriching the theoretical foundation of mechanistic interpretability. Here, we illustrate this alignment with several key examples:

\begin{enumerate}
  \item \textbf{Interpreting Polysemantic Neurons via Intervention Algebras.} Polysemantic neurons \cite{Scherlis2022PolysemanticityAC}, which encode multiple concepts, complicate interpretability. Using Intervention Algebras from causal abstraction, researchers can decompose neural networks into modular, interpretable components (like specific neuron groups or attention heads).
  \item \textbf{Graded Faithfulness through Approximate Abstraction}: Faithfulness \cite{jacovi2020faithfullyinterpretablenlpsystems}, or how accurately an explanation reflects a model's reasoning, should vary by context. Approximate Transformation, rooted in causal abstraction, provides flexible metrics for graded faithfulness.
  \item \textbf{Feature Attribution via Causal Abstraction.} Feature attribution methods, like LIME \cite{ribeiro2016whyitrustyou} and Integrated Gradients \cite{sundararajan2017axiomaticattributiondeepnetworks}, quantify feature impact on model behavior. Reinterpreted through causal abstraction, LIME approximates local causal fidelity, while Integrated Gradients can compute causal effects. 
  \item \textbf{Modular Feature Learning as Bijective Transformation.} Modular feature learning methods, such as PCA \cite{shlens2014tutorialprincipalcomponentanalysis} and Sparse Autoencoders \cite{cunningham2023sparseautoencodershighlyinterpretable}, which map high-dimensional activations to interpretable features, can be seen as using bijective transformations.
\end{enumerate}

\section{Dissucssion and Conclusion}

This compact survey highlights the latest advancements in \textbf{causal abstraction} theory and its applications for model interpretability. Causal abstraction is increasingly important in addressing mechanistic interpretability, especially for large-scale systems where understanding exact causal drivers behind behaviors like hallucination or alignment issues \cite{huang2023surveysafetytrustworthinesslarge,bereska2024mechanisticinterpretabilityaisafety} is critical. Previous methods, such as neuron- and representation-level analyses \cite{foote2023neurongraphinterpretinglanguage,zou2023representationengineeringtopdownapproach}, while effective, are largely experimental modeling techniques. For complex models like modern large language models, a foundational theoretical approach is needed, much like automaton extraction \cite{weiss2020,wei2024weighted,zhang2024automata}, which offered robust theoretical grounding but not always applicable for nowadays models.

Causal abstraction addresses this need by supporting counterfactual reasoning and causal analysis, as evidenced in scaled applications like Alpaca 7B \cite{10.5555/3666122.3669540}. However, the theory remains incomplete, largely centered on equivalence rather than true abstraction, and primarily formalizes existing methods. Future research could focus on (1) developing refined abstraction methods, as seen in DAS; (2) establishing exact causal dependencies between internal and external model behaviors; and (3) creating a streamlined theoretical framework tailored specifically to interpretability, minimizing broader causal theory complexities.

\newpage
\bibliographystyle{plain}
\bibliography{reason1}

\end{document}